\documentclass[lettersize,journal]{IEEEtran}
\usepackage{amsmath,amsfonts}
\usepackage{algorithmic}
\usepackage{algorithm}
\usepackage{array}
\usepackage[caption=false,font=normalsize,labelfont=sf,textfont=sf]{subfig}
\usepackage{textcomp}
\usepackage{stfloats}
\usepackage{url}
\usepackage{verbatim}
\usepackage{graphicx}
\usepackage{cite}
\usepackage{color}
\usepackage{makecell}
\usepackage{hyperref}
\hypersetup{
    colorlinks=true,
    urlcolor=magenta
}

\begin{document}

\title{Bootstrapping Interactive Image-Text Alignment for Remote Sensing Image Captioning}

\author{Cong Yang, Zuchao Li, Lefei Zhang,~\IEEEmembership{Senior Member}
\thanks{Manuscript received 23 August 2023, revised 2 December 2023. This work was supported by the National Natural Science Foundation of China under Grant 62122060 and the Special Fund of Hubei Luojia Laboratory under Grant 220100014. (Corresponding author: Lefei Zhang.)}
\thanks{The authors are with the National Engineering Research Center for Multimedia Software, School of Computer Science, Wuhan University, Wuhan, 430072, P. R. China, and also with the Hubei Luojia Laboratory, Wuhan 430079, P. R. China. (e-mail: \{yangcong356, zcli-charlie, zhanglefei\}@whu.edu.cn)}}

\markboth{Journal of \LaTeX\ Class Files,~Vol.~14, No.~8, August~2021}%
{Shell \MakeLowercase{\textit{et al.}}: Vision-Language Pre-training for Remote Sensing Image Captioning}


\maketitle

\begin{abstract}
Recently, remote sensing image captioning has gained significant attention in the remote sensing community. Due to the significant differences in spatial resolution of remote sensing images, existing methods in this field have predominantly concentrated on the fine-grained extraction of remote sensing image features, but they cannot effectively handle the semantic consistency between visual features and textual features. To efficiently align the image-text, we propose a novel two-stage vision-language pre-training-based approach to bootstrap interactive image-text alignment for remote sensing image captioning, called BITA, which relies on the design of a lightweight interactive Fourier Transformer to better align remote sensing image-text features. The Fourier layer in the interactive Fourier Transformer is capable of extracting multi-scale features of remote sensing images in the frequency domain, thereby reducing the redundancy of remote sensing visual features. Specifically, the first stage involves preliminary alignment through image-text contrastive learning, which aligns the learned multi-scale remote sensing features from the interactive Fourier Transformer with textual features. In the second stage, the interactive Fourier Transformer connects the frozen image encoder with a large language model. Then, prefix causal language modeling is utilized to guide the text generation process using visual features. Ultimately, across the UCM-caption, RSICD, and NWPU-caption datasets, the experimental results clearly demonstrate that BITA outperforms other advanced comparative approaches. The code is available at \href{https://github.com/yangcong356/BITA} {https://github.com/yangcong356/BITA}.

\end{abstract}


\section{Introduction}
\IEEEPARstart{R}{emote} sensing image captioning (RSIC) is a complex task that revolves around enabling machine learning models to accurately comprehend and depict various contextual objects within remote sensing images using appropriate vocabulary\cite{WHLY2023}. In contrast to other remote sensing tasks such as image classification \cite{HGYY2021, DLZD2021, RDHR2023, YHWL2023, ZZD2016, ZZ2022} or object detection \cite{LLYD2022, SWWL2021}, RSIC stands apart; its objective is not merely to predict isolated labels or words, but rather to generate meaningful and comprehensive sentences that describe the entirety of a remote sensing image. Due to its significant potential applications in areas like smart city development and military intelligence \cite{SZ2017}, RSIC is becoming increasingly appealing within the field of remote sensing. 

Recent works \cite{YNR2022, HCMS2022, LZHW2021} have yielded significant improvements in captioning performance on the RSIC task. Zhang et al. \cite{ZLAG2017} employed convolutional neural networks to detect major objects in remote sensing images, subsequently utilizing a recurrent neural network to generate natural language descriptions for the initially detected targets. Wang et al. \cite{WLZL2019} utilized the Mahalanobis matrix to measure the difference between visual features and textual features to improve captioning results. Due to prior work neglecting the structured spatial relations of semantic content in remote sensing images, Zhao et al. \cite{ZSZ2022} proposed a structured attention-based approach that employs pixel-level segmentation masks to guide the caption generation process, enabling better generation of accurate and concise descriptions. 
While these advanced methods have improved the accuracy of caption generation, they overlook bridging the modality gap between visual and textual information, resulting in generated descriptive sentences lacking semantic coherence \cite{WHLY2023}. The modality gap is manifested as the image-text alignment issue, which involves matching specific objects within remote sensing images to their corresponding proper nouns in the generated sentences.

Due to the capability of reducing the modality gap, the vision-language pre-training (VLP) paradigm has gained increasing attention in multimodal understanding and generation tasks \cite{YLWY2021, CGYL2022, WYYD2022}. The goal of VLP is to integrate the robust representations and reasoning capabilities of both visual models and large language models (LLMs, such as GPT-3\cite{BMRS2020}, T5\cite{RSRL2020}, OPT\cite{ZRGA2022}, LLaMA\cite{TLIM2023}, etc.), enabling effective learning from multimodal data. Although the clear capability of the VLP paradigm to bridge multi-modal discrepancies is undeniable, the multi-scale characteristic of remote sensing images presents a significant challenge in the development path for VLP in the remote sensing field. This is because the multi-scale characteristic of remote sensing images primarily manifests as significant variance in the ground sampling distance (reflecting the spatial resolution of the remote sensing sensor) among the images. This variance can affect the feature representations of the same object, thereby increasing the modality gap between visual information and textual information \cite{WZZG2022}.

To address the negative effects caused by the multi-scale characteristics of remote sensing images and the multimodal gap between image and text, this paper proposes an interactive visual-textual alignment method for remote sensing image captioning, named BITA. a novel lightweight Interactive Fourier Transformer (IFT) module is designed to guide the alignment between images and texts during a two-stage pre-training process. The IFT module represents a lightweight encoder-only Transformer that makes use of a set of trainable embeddings for extracting valuable visual features from the frozen image encoder. Moreover, the IFT module serves as an information-filtering component between the frozen image encoder and the frozen LLM. It accomplishes this by learning image-text alignment through contrastive learning, enabling it to understand the matching relationships between image-text pairs. The learned useful visual features, then, are selected as visual prompt inputs for the LLM, thus guiding the language model to produce the description. In the first pre-training stage, we employ image-text contrastive learning \cite{AJCA2021} to constrain the IFT module to learn the most relevant and valuable visual representations that are closely associated with the text. In the second pre-training stage, we concatenate the visual features learned by IFT with the encoded textual features. This combined representation is then input into the LLM, utilizing language modeling learning \cite{DYWW2019} to guide the generative learning of visual-to-language. Throughout this process, we ensure that IFT remains trainable, allowing the whole model to establish a causal reasoning relationship between visual and textual features.

Overall, the main contributions of this paper can be summarized in the following three aspects:
\begin{enumerate}
    \item {We introduce the VLP paradigm into the RSIC task and propose a novel VLP model specifically designed for RSIC. By utilizing the first-stage pre-training process constrained by image-text contrastive learning and the second-stage pre-training process guided by language modeling, BITA is capable of acquiring robust visual features, achieving visual-semantic alignment of objects in remote sensing image-text pairs.}
    \item {We devise an IFT module, which acts as an intermediary between the frozen visual encoder and the frozen LLM. This module employs a parameter-free Fourier transform to encode image and text information, reducing the model's parameter. Moreover, it efficiently learns the multi-scale features of remote sensing images in the frequency domain.}
\end{enumerate}

\section{Related Work}

\subsection{Remote Sensing Image Captioning}
The current prevailing methods for RSIC all adopt a unified encoder-decoder framework\cite{ZSZ2022, WHZL2021, HWL2021, LZHW2021, SND2021, LWZ2020}. This framework typically uses convolutional neural networks like ResNet, VGG, and GoogleNet as encoders to extract deep semantic visual features from remote sensing images. Meanwhile, decoders often use recurrent neural networks to translate these features into corresponding sentence descriptions. Furthermore, there are also a few studies that utilize Transformers \cite{LZS2022, KSBZ2022} and support vector machines (SVM) \cite{HM2022} as decoders in RSIC tasks. 

Shi et al. \cite{SZ2017} were the pioneers in investigating image captioning in the field of remote sensing. They pointed out that the main challenge in RSIC is that models need to not only capture a variety of objects at different scales but also express their attributes and interaction states. Wang et al. \cite{WHZL2021} criticized the encoder-decoder architecture for lacking interpretability. Consequently, they introduced a word-sentence framework composed of a word generator and a sentence generator. The former extracted attribute information from significant objects in remote sensing images, and the latter then assembled this information into syntactically correct and coherent sentences. 
Considering the presence of multi-scale characteristics in remote sensing images, Wang et al. \cite{WZZG2022} introduced a two-stage multi-scale structural representation method. This approach boosted the model's object differentiation ability through multi-level CNN feature interactions, increasing description accuracy. Li et al. \cite{LZGL2022} proposed a recurrent attention and semantic gate (RASG) framework to interact with both image content and sentence, which is devised to extract and understand effective information from both the complex content of remote sensing images and textual information. 

Due to the remarkable success of attention mechanisms in both the CV and NLP domains, there have been plenty of attention-based methods in the field of RSIC as well. Wang et al. \cite{WHZL2022} proposed an attention-based global-local captioning model (GLCM) to obtain global-local visual feature representation. Additionally, GLCM employed a similarity-based approach to measure the relationships between all generated words and their associations with the most relevant local visual features. Cheng et al. \cite{CHXZ2022} proposed a multilevel and contextual attention network (MLCA-Net) that adaptively aggregates image features from specific spatial regions and scales, while also incorporating a contextual attention module to explore latent context. Zhang et al. \cite{ZZYG2022} proposed a global visual feature-guided attention (GVFGA) mechanism and a linguistic state-guided attention (LSGA) mechanism for RSIC. The former was utilized to filter out redundant feature components from the fused image features, ensuring more prominent visual features. The latter enhanced the fusion of visual and textual features and eliminated irrelevant information.

However, these encoder-decoder-based approaches primarily leverage attention mechanisms and multi-scale aggregation modules to unearth complex spatial relationships and multi-scale information of objects within remote sensing images, without considering whether the alignment between image and text is achieved.

\subsection{Vision-Language Pre-training}
Recently, the VLP paradigm has gained remarkable performance in vision-language understanding and generation tasks, such as visual question answering (VQA) \cite{ZPZH2020} and natural image captioning (NIP) \cite{BCCB2022}, etc. Based on whether the model can be pre-trained in an end-to-end manner, VLP can be categorized into two approaches: end-to-end pre-training and two-stage pre-training.

The end-to-end pre-training approach is an earlier VLP paradigm. In this kind of method, the initial step involves using a pre-trained object detector (such as a pre-trained Faster-RCNN model \cite{AHBT2018}) to capture image region features, which are then fed into a cross-modal decoder to generate image captions. ViLBERT \cite{LBPL2019} and LXMERT \cite{TB2019} both extracted image region features and subsequently employed a co-attentional Transformer extended from the popular BERT \cite{DCLT2019} architecture for multimodal fusion. Unicoder-VL \cite{LDFG2020} employed a merged attention fusion module, which inputs both image region and text features into a universal Transformer encoder. Moreover, masked language modeling, masked object classification, and visual-linguistic matching pre-training tasks were utilized to learn context-aware representations. The crucial role of visual features for image-text semantic alignments was showcased by Oscar \cite{LYLZ2020} and VinVL \cite{ZLHY2021}: Oscar used detected image object tags as an anchor point to simplify alignment learning, while VinVL employed a more robust object detector based on Oscar for feature extraction.

The two-stage pre-training approach is generally an end-to-end method that utilizes either CNNs or a vision Transformer \cite{DBKW2021} to extract image grid/patch embeddings. \cite{HZLF2020} proposed Pixel-BERT, a unified end-to-end framework with a CNN-based visual encoder and multimodal Transformer, which is capable of aligning semantic connections between image pixels and text. 
PureT \cite{WXS2022} utilized a Swin-Transformer to extract grid-level features and captured the intra-relationship between image and text by adding a pre-fusion process. Due to the robust representation and inference ability of visual foundation models (VFMs) and LLMs, Flamingo \cite{ADLM2022},  leveraging large pre-trained vision-only and language-only models, achieved few-shot learning capabilities by aligning arbitrary interleaved sequences of visual and textual data. The primary challenge when employing a fixed LLM is to align visual features with the textual context \cite{LLSH2023}. BLIP-2 has demonstrated exceptional performance \cite{LLSH2023} for vision-language understanding and generation tasks. Similar to Flamingo, BLIP-2 adopted a pre-trained frozen image encoder and a frozen LLM to initiate the visual-language pre-training process. Additionally, BLIP-2 also constructed a query Transformer (Q-Former) to bridge the modality gap, and this module plays a role in both stages of the two-stage pre-training process. Nevertheless, the Q-Former is completely based on BERT, which causes the Q-Former to take more time when interacting with image-text information. Moreover, bidirectional attention in BERT suffers from a low-rank issue that can weaken the model's expressive capacity \cite{DCL2021}, a drawback that Q-Former is likely to inherit.

\section{Methodolgy}
Prior to delving into the method proposed in this article, it's essential to outline the foundational framework of VLP and the Fourier transform that makes up the IFT module. We, then, provide a detailed description of a VLP method proposed for the RSIC task, known as BITA, which harvests the powerful representation and reasoning capabilities of frozen pre-trained VFMs and LLMs. To bridge the modality gap between multimodal data and simultaneously capture multi-scale features within remote sensing images, we propose a novel IFT module in the two-stage pre-training process for image-text alignment: (1) aligning image-text by representation and (2) visual feature-guided language generative learning.


\subsection{Vision-Language Pre-Training Setup}
The VLP involves utilizing deep neural networks to extract image and text features from pre-training dataset with $N$ pairs of image-text, denoted as $\mathcal{D} = \left \lbrace \mathbf{x}_{n}^{g}, \mathbf{x}_{n}^{t} \right \rbrace_{n=1}^{N}$. Here, $\mathbf{x}_{n}^{g}$ and $\mathbf{x}_{n}^{t}$ represent image and text samples respectively, forming pairs. The deep neural networks consist of an image encoder $f_{\theta}$ and a text encoder $f_{\phi}$.  These encoders are responsible for encoding the image and text (from the image-text pairs $\left \lbrace \mathbf{x}_{n}^{g}, \mathbf{x}_{n}^{t} \right \rbrace_{n=1}^{N}$) into respective embeddings: $\mathbf{z}_{n}^{g} = f_{\theta}\left( \mathbf{x}_{n}^{g} \right)$ for image embedding and $\mathbf{z}_{n}^{t} = f_{\phi}\left( \mathbf{x}_{n}^{t} \right)$ for text embedding. For the image captioning task, the image and text embedding are aligned through a predefined pre-training task. These learned visual-textual semantic representations are then fed into a decoder $f_{\psi}$ to generate the desired text output $y$:

\begin{equation}\label{eq1}
    y = f_{\psi}\left( \left \lbrack f_{\theta}\left( \mathbf{x}_{n}^{g} \right), f_{\phi}\left( \mathbf{x}_{n}^{t} \right) \right \rbrack \right),
\end{equation}

\noindent where $\left \lbrack \cdots \right \rbrack$ represents tensor concatenation.

\subsection{Discrete Fourier Transform}
The most crucial component in building the IFT module is the Fourier transform, which plays a significant role in digital signal processing. For the sake of simplicity, let's start by introducing the 1-dimensional discrete Fourier transform (1D DFT). Given a sequence $\left \lbrace \mathbf{x}_{m} \right \rbrace$ of length $m \in \left[0, M-1 \right]$, the formula for the 1D DFT that transforms this sequence into the frequency domain can be represented as follows:

\begin{equation}\label{eq2}
\mathbf{x}^{\prime}_{k}=\sum_{m=0}^{M-1} e^{-i \left(\frac{2 \pi}{M}\right) m k} \mathbf{x}_m, \quad 0 \leq k \leq M-1,
\end{equation}

\noindent for each $k \in \left[0, M-1 \right]$, the DFT generates a new representation $\mathbf{x}^{\prime}_{k}$, which is the sum of all original input tokens $\mathbf{x}_m$. DFT finds widespread application in signal processing algorithms for two main reasons: (1) Both the input and output of DFT are discrete, making them computationally manageable for computers; (2) Efficient algorithms exist for computing DFT, namely the Fast Fourier Transform (FFT) \cite{CT1965}. FFT algorithm leverages the symmetry and periodicity of DFT matrix $W_{km}=\left(e^{-i \left(\frac{2 \pi}{M}\right) m k} / \sqrt{M} \right)$ to reduce the complexity of computing the DFT from $\mathcal{O}\left( M^{2} \right)$ to $\mathcal{O}\left( M \log M \right)$ on graphics processing units (GPUs) \cite{RZZL2021}. Additionally, the DFT can be extended to handle 2D signals as well. The 2D DFT can be thought of as performing alternating 1D DFT along the two dimensions of a 2D sequence. Since the input 2D sequence exhibits conjugate symmetry, the FFT algorithm can also be applied to 2D DFT to enhance computational efficiency.

\begin{figure*}[t]
\centering
\includegraphics[width=0.75\linewidth]{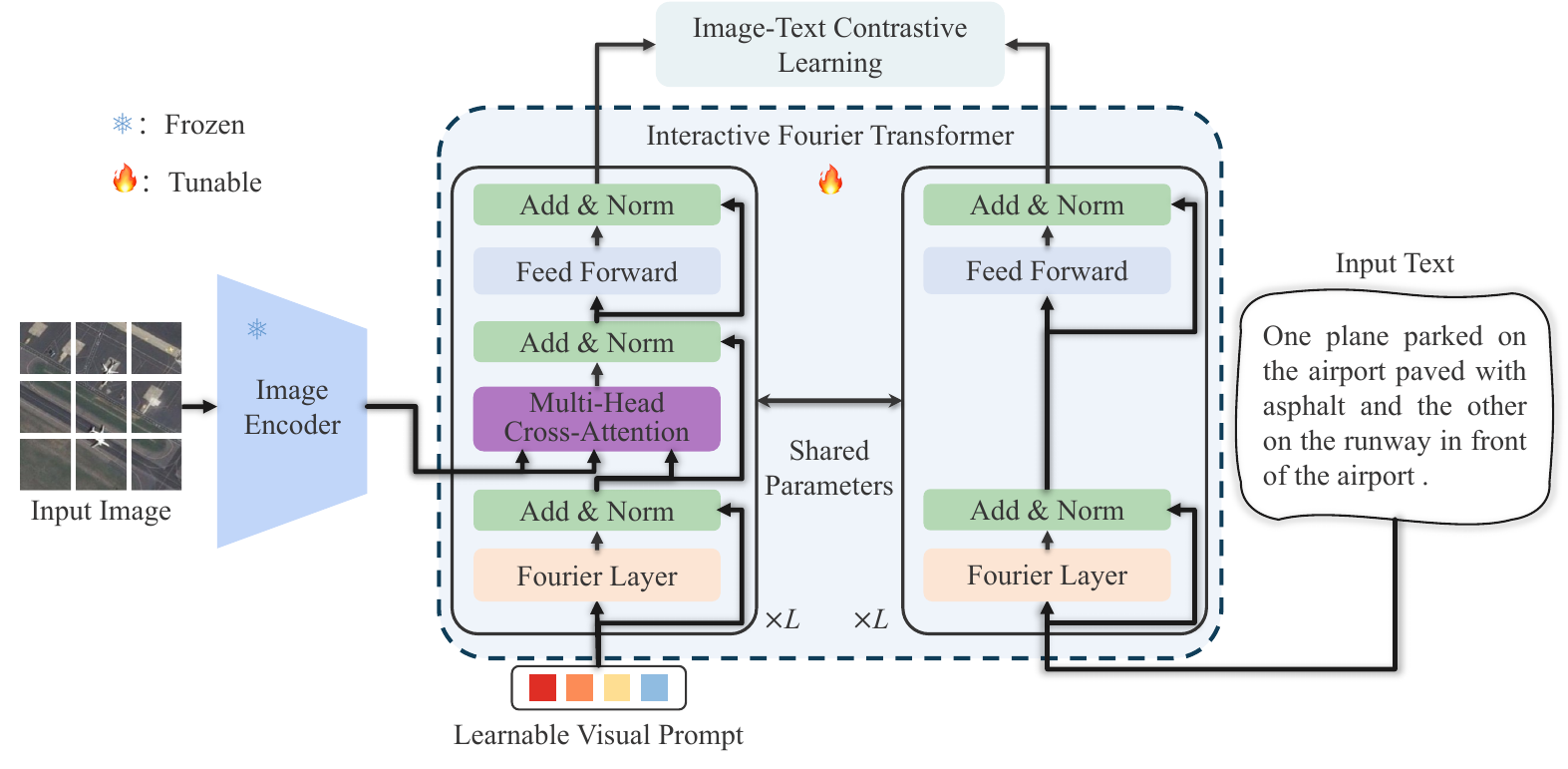}
\caption{The overall framework for the first stage of BITA's image-text representation learning. The learnable visual prompts are input into the Fourier-based image Transformer (left side of the Interactive Fourier Transformer) to interact with the features obtained from the frozen image encoder, capturing the most useful visual features. The original textual information is input into the Fourier-based text Transformer (right side of the Interactive Fourier Transformer) for text feature extraction. $L$ represents the number of layers in these Transformers. Finally, the extracted most useful visual features and textual features undergo image-text contrastive learning. The image-text contrastive learning is to minimize the distance between positive image-text pairs and increase the distance between negative image-text pairs, to achieve alignment of image-text features.}
\label{ch3-fig-1}
\end{figure*}

\subsection{Model Architecture}

Replacing the self-attention layers in the Transformer with a standard Fourier transform can help reduce the number of network parameters, while still maintaining model performance approximately unchanged \cite{JJIS2022, RZZL2021}. Furthermore, the Fourier transform can decompose the input image into different frequency components, which can reflect the multi-scale features of the input image \cite{CYC2023, XECZ2022}. Inspired by these, we construct a lightweight and trainable IFT module using a parameter-free Fourier transform and a cross-attention mechanism. The IFT facilitates image-text alignment between the frozen image encoder and the frozen LLM. As shown in Fig. \ref{ch3-fig-1}, the IFT comprises two sub-modules with shared parameters: Fourier-based image Transformer $f_{\delta}$ and Fourier-based text Transformer $f_{\phi}$. The former module is used to interact with the features $\mathbf{z}_{n}^{g}$ obtained from the image encoder and the learnable visual prompt $\mathbf{z}_{v}$, while the latter module is employed to process the raw textual information $\mathbf{x}_{n}^{t}$. In both of these sub-modules, the most critical aspect is the construction of the Fourier layer. Specifically, the Fourier layer employs a 2D DFT to process input embedding, with one 1D DFT operating along the sequence length dimension $\mathcal{F}_{seql}$ and another 1D DFT operating along the hidden dimension $\mathcal{F}_{h}$:

\begin{equation}\label{eq3}
\begin{aligned}
    {\mathbf{z}_{v}}^{\prime} = \mathcal{F}_{seql}\left(\mathcal{F}_{h}\left(\mathbf{z}_{v}\right)\right),\\
    {\mathbf{z}^{t}}^{\prime} = \mathcal{F}_{seql}\left(\mathcal{F}_{h}\left(\mathbf{x}_{n}^{t}\right)\right),
\end{aligned}
\end{equation}

\noindent where ${\mathbf{z}_{v}}^{\prime}$ and ${z^{t}}^{\prime}$ represent the features after processing through the Fourier layer, respectively. Notably, to avoid altering other components of the model, we retain only the real part of the Fourier-transformed outputs, comprising all frequency information of the signal. As image and text data are discrete, sparse, and non-periodic, phase information is not as crucial. Furthermore, the model incorporates additional linear layers and residual connections, which can mitigate the potential loss caused by discarding phase information \cite{JJIS2022}.

For the Fourier-based image Transformer, we generate 32 learnable visual prompt embeddings, serving as inputs to the Fourier-based image Transformer. This configuration allows the Fourier-based image Transformer module to extract a consistent number of output features from the image encoder, regardless of the initial input image's resolution. The visual prompt embeddings ${\mathbf{z}_{v}}^{\prime}$ after the Fourier layer can interact with the visual features extracted from the frozen image encoder through cross-attention (CA) layers, achieving an efficient lower-dimensional representation of visual features. This process can be expressed by the following formula:

\begin{equation}\label{eq4}
\begin{aligned}
    \mathbf{Q} = {\mathbf{z}_{v}}^{\prime} \mathbf{W}^Q,     \mathbf{K} &= \mathbf{z}_{n}^{g} \mathbf{W}^K, \mathbf{V} = \mathbf{z}_{n}^{g} \mathbf{W}^V, \\
    \text{CA}(\mathbf{Q}, \mathbf{K}, \mathbf{V}) &= \text{softmax}\left(\frac{\mathbf{Q}\mathbf{K}^T}{\sqrt{d_k}}\right)\mathbf{V},
\end{aligned}
\end{equation}

\begin{equation}\label{eq5}
    \begin{aligned}
        \text{MultiHead}(\mathbf{Q}, \mathbf{K}, \mathbf{V}) &= \text{Concat}(\text{head}_1, ... , \text{head}_h)\mathbf{W}^O, \\
        \text{where}\ \text{head}_j &= \text{CA}(\mathbf{Q}_j, \mathbf{K}_j, \mathbf{V}_j),
    \end{aligned}
\end{equation}

\noindent where $\mathbf{Q}, \mathbf{K}, \text{and} \mathbf{V}$ represent Query, Key, and Value, respectively, all obtained through linear mappings from relevant features. $\sqrt{d_k}$ is the dimension of $\mathbf{K}$, used to scale the dot product to prevent it from becoming too large in high dimensions. The softmax function is applied to each row to normalize the weights into a valid probability distribution. In the multi-head cross-attention mechanism (with $h$ being the number of heads), cross-attention operations are independently and parallelly executed multiple times, and the final output is concatenated ($\text{Concat}$) and obtained as the final output through an additional linear mapping $\mathbf{W}^O$.

For the Fourier-based text Transformer, we first transform the raw textual information into embeddings of text tokens. Due to the Fourier layer's role as a token mixing mechanism \cite{JJIS2022}, we then utilize the Fourier layer to capture dependencies among these tokens. Additionally, the shared feed-forward layers can serve as an interaction mechanism between the learnable visual prompt embeddings and the text token embeddings. The process of the Fourier-based text Transformer handling raw textual information can be represented as:

\begin{equation}\label{eq6}
    \mathbf{z}_{n}^{t} = f_{\phi}\left( \mathbf{x}_{n}^{t} \right),
\end{equation}

\noindent where $\mathbf{z}_{n}^{t}$ represents the output features of the Fourier-based text Transformer.

To bridge the gap between images and texts, we employ the image-text contrastive learning pre-training task \cite{AJCA2021} in the first pre-training stage to involve the learned visual prompt and text embeddings, which enforce alignment between images and texts. Except for the Fourier layer, the other network structure parameters of the two sub-modules in the IFT are consistent with the parameters of BERT-base \cite{DCLT2019}. Therefore, except for the cross-attention layers that are initialized randomly, the rest of the trainable components in the IFT module utilize pre-trained weights from BERT-base. The IFT module encompasses a total of 171 million parameters.

\begin{figure*}[!t]
\centering
\includegraphics[width=0.7\linewidth]{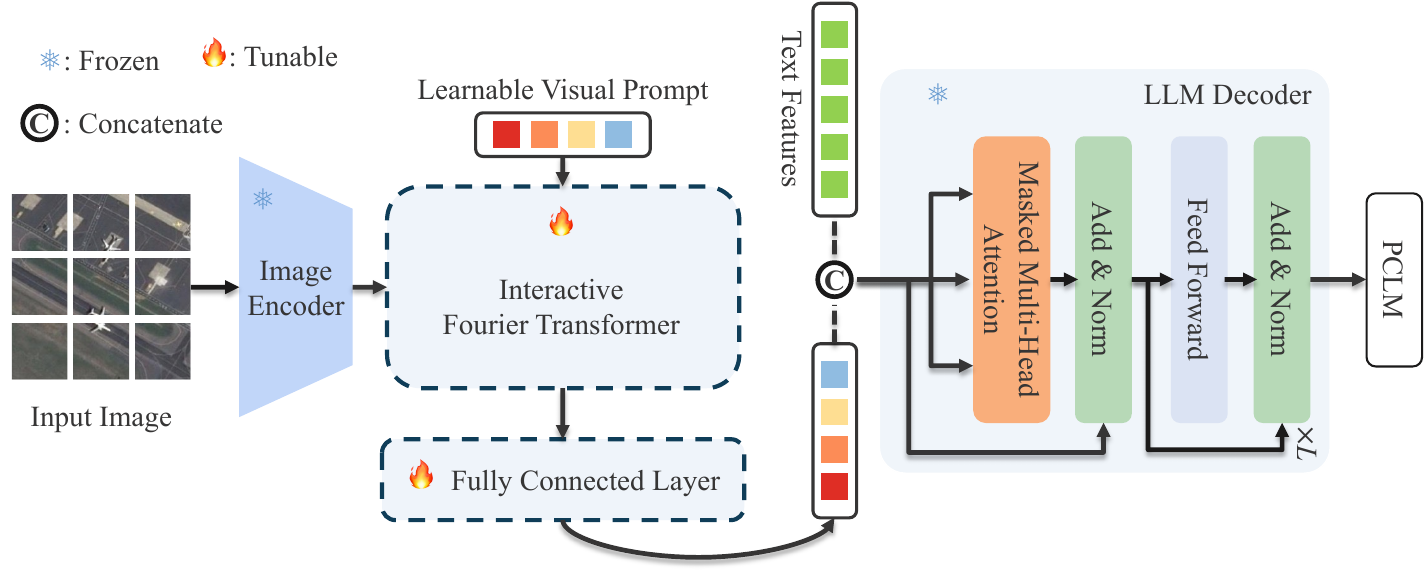}
\caption{The overall framework for the second stage of BITA's visual feature-guided language generative learning. Learned visual prompt serves as prefixes for text features, which are then inputted into a large language model for controlled text generation. $L$ represents the number of layers in the LLM decoder, where the masked multi-head attention mechanism \cite{VSPU2017} is employed to mask out sequence information beyond the current time step, enhancing the model's inferential capabilities.}
\label{ch3-fig-3}
\end{figure*}

\begin{figure}[t]
\centering
    \includegraphics[width=0.8\linewidth]{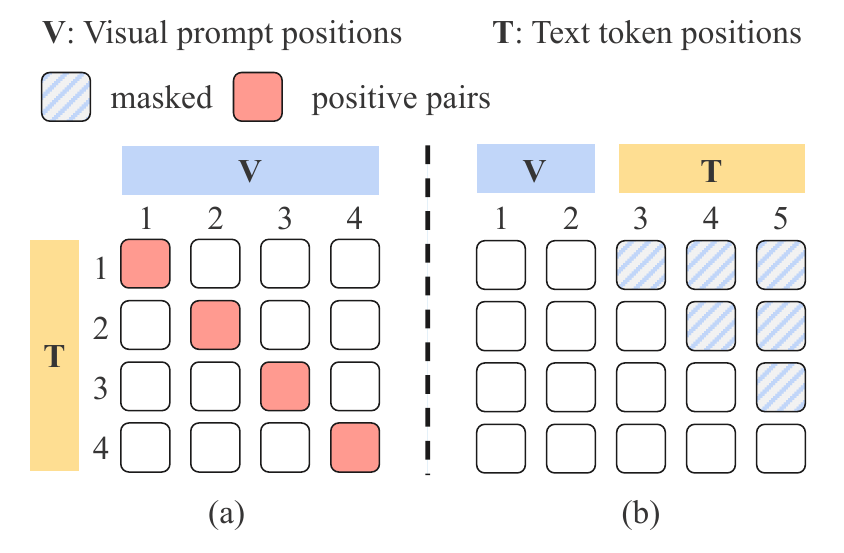}
\caption{An illustration of Image-Text Contrastive Learning and Prefix Causal Language Modeling: (a) represents Image-Text Contrastive Learning, where the pink squares represent positive image-text pairs, and the blank squares represent negative image-text pairs; (b) represents Prefix Causal Language Modeling, where the masked squares represent masked text tokens that can attend to all visual prompts and the preceding text tokens.}
\label{ch3-fig-2}
\end{figure}

\subsection{Aligning Image-Text by Representation Learning}
In the representation learning phase, we connect the IFT module to the frozen image encoder and perform pre-training using image-text pairs. Our objective is to train the IFT module so that the visual prompts can learn to extract visual representations most relevant to the text. Inspired by approaches like BLIP-2 \cite{LLSH2023} and CLIP \cite{AJCA2021}, we find that optimizing solely the image-text contrastive learning pre-training objective can yield strong performance. This stands in contrast to BLIP-2, which employs three different training objectives.

\textbf{Image-Text Contrastive Learning (ITC)} aims to maximize the mutual information between images and texts to learn a joint representation. ITC achieves this by contrasting the similarity of positive image-text pairs against the similarity of negative pairs in a batch $\mathcal{B}$. We align the output visual prompt embeddings $\hat{\mathbf{z}}_{v}$ from the Fourier-based image Transformer with the representation $\mathbf{z}_{\text{CLS}}^{t}$ from the Fourier-based text Transformer, where $\mathbf{z}_{\text{CLS}}^{t}$ corresponds to the embedding of the $\left[\text{CLS}\right]$ token. When computing the ITC loss, we employ a bi-directional objective for image-text pairs:

\begin{equation}\label{eq7}
    \mathcal{L}_{\text{ITC}} = \mathcal{L}_{g2t} + \mathcal{L}_{t2g}.
\end{equation}

In Eq. \ref{eq8}, the contrastive loss from images to texts $\mathcal{L}_{g2t}$ aligns the matched images in the batch with the given texts, which can be written as:

\begin{equation}\label{eq8}
    \mathcal{L}_{g 2 t}= -\frac{1}{\mathcal{B}} \sum_{k \in \mathcal{P}(u)} \log \frac{\exp \left({\hat{\mathbf{z}}_{v_u}}^{T} {\mathbf{z}_{\text{CLS}_{k}}^{t}} / \tau\right)}{\sum_{q=0}^{\mathcal{B}} \exp \left({\hat{\mathbf{z}}_{v_u}}^{T} {\mathbf{z}_{\text{CLS}_{q}}^{t}} / \tau \right)}
\end{equation}

\noindent where $k \in \mathcal{P}(u)=\left\{k \mid k \in \mathcal{B}, \mathbf{x}_k^t=\mathbf{x}_u^g\right\}$.
$\tau$ is the temperature of the softmax for the normalization, and we set it to 0.07.

In Eq. \ref{eq9}, the contrastive loss from texts to images $\mathcal{L}_{t2g}$ aligns the matched texts in the batch with the given images, which can be written as:

\begin{equation}\label{eq9}
    \mathcal{L}_{t 2 g}= -\frac{1}{\mathcal{B}} \sum_{k \in \mathcal{P}(q)} \log \frac{\exp \left(\left(\mathbf{z}_{\text{CLS}_{q}}^{t}\right)^{T} \hat{\mathbf{z}}_{v_k} / \tau \right)}{\sum_{u=0}^{\mathcal{B}} \exp \left(\left(\mathbf{z}_{\text{CLS}_{q}}^{t}\right)^{T} \hat{\mathbf{z}}_{v_u} / \tau \right)}
\end{equation}

\noindent where $k \in \mathcal{P}(q)=\left\{k \mid k \in \mathcal{B}, \mathbf{x}_k^g=\mathbf{x}_q^t\right\}$. The value of $\tau$ is set to be the same as in Eq. \ref{eq8}. Eventually, we use Fig. \ref{ch3-fig-2}(a) to visually illustrate the computational concept of bi-directional ITC.

\subsection{Visual Feature-Guided Language Generative Learning}
During the visual feature-guided language generative pre-training phase, we connect the pre-trained IFT (along with the frozen image encoder) from the representation learning stage to the frozen LLM to leverage the language generation and reasoning capabilities of the LLM. As shown in Fig. \ref{ch3-fig-3}, we employ a fully connected layer to linearly project the output visual prompt embeddings $\hat{\mathbf{z}}_{v}$ to the same dimension as the input text embeddings $\mathbf{z}_{n}^{t}$ for the LLM. To enable the LLM to generate controlled text outputs conditioned on the visual prompts extracted by the IFT, we concatenate the projected visual prompt embeddings as a prefix to the input text embeddings. We utilize a prefix causal language modeling to control the interaction between visual prompt embeddings and text embeddings. Since the IFT has been pre-trained, it can extract visual representations that are relevant to text features. This enables it to effectively serve as an information bottleneck, conveying the most useful information to the LLM while filtering out visually irrelevant details. 

\textbf{Prefix Causal Language Modeling (PCLM)} is a kind of language modeling that aims to generate text by conditioning on a given image prefix, similar to the one used in UniLM \cite{DYWW2019}. PCLM is similar to traditional causal language modeling \cite{LTZN2022}, where the model generates text one token at a time, conditioned on the previous tokens. Due to the structure of the IFT, direct interaction between the frozen image encoder and the text tokens is not allowed. Therefore, it is essential to utilize the PCLM, guiding text generative process by visual features. As shown in Fig. \ref{ch3-fig-2}(b), each text token embedding can attend to all visual prompt embeddings and its previous text token embeddings. Note that the visual prompt embeddings added as prefixes do not contribute to the computation of the PCLM loss which is still essentially a sequence-to-sequence language modeling loss.

\section{Experiments}
\subsection{Datasets Description}
This subsection introduces three remote sensing image captioning datasets, providing detailed descriptions based on data volume and textual semantic richness. Performing remote sensing image captioning tasks on the NWPU-caption dataset is the most challenging among the three datasets. The detailed descriptions are as follows:
\begin{enumerate}
    \item {\textit{The UCM-caption dataset} \cite{QLTL2016}, is an extension of the UC Merced Land Use dataset, which comprises 2,100 aerial remote sensing images. The image resolution of each image is $256 \times 256$, with a ground sampling distance of 0.3 meters. The RGB three-band image dataset was collected from the USGS National Map Urban Area Imagery. Each image in the UCM-caption dataset is annotated with five unique sentences, resulting in a total of 10,500 annotations.}
    \item {\textit{The RSICD dataset} \cite{LWZL2018} is a large-scale aerial remote sensing image captioning dataset, containing 10,921 remote sensing RGB three-band images collected from Baidu Map, MapABC, Google Earth, and Tianditu. Each image has a fixed size of $224 \times 224$ but varies from ground sample distance. The RSICD dataset comprises a total of 24,333 sentences. The annotations are detailed as follows: 724 images are accompanied by five distinct sentences; 1,495 images have descriptions in four varying sentences; 2,182 images are matched with three diverse sentences; 1,667 images are matched with two unique sentences, and 4,853 images have a single sentence. RSICD increased sentence diversity by expanding the dataset to 54,605 sentences through random duplication of existing sentences when fewer than five distinct sentences were available for the same image.}
    \item {\textit{The NWPU-caption dataset} \cite{CHXZ2022}, is built upon the remote sensing image classification dataset created by Northwestern Polytechnical University, incorporating 31,500 aerial remote sensing images with RGB three-band data and 157,500 sentences for remote sensing image caption. Each image has a spatial resolution of $256 \times 256$ pixels, with ground sampling distances ranging from 0.2 to 30 meters. Additionally, to enhance the semantic richness of sentences, each image is accompanied by five distinct textual descriptions.}
\end{enumerate}

\subsection{Model Pre-Training Setup}
\subsubsection{Pre-Training Dataset}
Similar to BLIP-2, in the pre-training stage, we combine the UCM-caption, RSICD, and NWPU-caption datasets for the two-stage pre-training. The total number of training images is 57,933, which amounts to 289,665 image-text pairs. After being augmented by random resized cropping and random horizontal flipping, the size of the images is $224 \times 224$.

\subsubsection{Frozen Image Encoder and LLM}
For the frozen image encoder, we explore the SOTA pre-training vision Transformer model ViT-L/14 from CLIP\footnote{\href{https://storage.googleapis.com/sfr-vision-language-research/LAVIS/models/BLIP2/clip_vit_L.pth}{Pre-trained weights for CLIP ViT-L/14}} \cite{AJCA2021}. To avoid excessive normalization from the final layer norm in pre-trained CLIP ViT-L/14, we remove this layer and utilize the output features from the second-to-last layer for image-text alignment. For the frozen language model, we explore the unsupervised trained decoder-only OPT 2.7B\footnote{\href{https://huggingface.co/facebook/opt-2.7b}{Pre-trained weights for OPT 2.7B}} language model \cite{ZRGA2022}, which is a high-performance, versatile open-source LLM. Additionally, the decoder-only OPT 2.7B language model is well-suited for prompt-guided language understanding and generation tasks \cite{TDTG2023, DSDH2023}.

\subsubsection{Pre-Training Hyperparameters}
The following hyperparameters are used for both image-text representation learning and language generative learning. We use AdamW optimizer with $\beta_{1}=0.9$, $\beta_{2}=0.98$, and a weight decay of 0.05. For the first 5000 training steps, we utilize linear learning rate warmup, with an initial learning rate set to $1e-6$. Upon reaching an initial learning rate of $1e-4$, we employ the cosine learning rate decay and set the minimum decaying learning rate to $1e-5$. We conduct pre-training for only 5 epochs in both the image-text representation learning and language generative learning stages. We employ automatic mixed precision and distributed data-parallel strategy for pre-training on four NVIDIA GeForce RTX 4090 GPUs. For the first-stage pre-training and second-stage pre-training, the batch sizes per card were 96 and 64, respectively.

\subsubsection{Fine-Tuning Hyperparameters}
After completing the two-stage pre-training tasks, we perform fine-tuning on a single dataset. We still employ the AdamW optimizer consistent with the pre-training tasks and set the same parameters. Similarly, we utilize linear learning rate warmup and cosine learning rate decay, but with different parameter settings compared to two-stage pre-training. We apply linear learning rate warmup for the first 2000 fine-tuning steps, with an initial learning rate of $1e-8$. Once the initial learning rate reaches $1e-5$, we proceed with cosine learning rate decay, setting the minimum decayed learning rate to $0$. Similarly, during the fine-tuning stage, we also employ automatic mixed precision and distributed data-parallel strategy. We set the batch sizes for the training and validation steps as 64 and 16, respectively. Moreover, we set the maximum length of the output sequences to the maximum token count in the dataset itself. Through analysis, the maximum token counts for UCM-caption, RSICD, and NWPU-caption datasets are 22, 34, and 50, respectively. To enhance the fault tolerance of caption generation, we utilize the heuristic search algorithm known as beam search. Based on prior research \cite{ZSZ2022} and common practices \cite{WXS2022, HGWY2022, CDW2022}, we set the beam size to 5 to balance accuracy and search efficiency. As a result, the model generates five candidate sentences and selects the best among them.

\subsection{Evaluation Metrics}
To assess the model's performance from various perspectives, we selected five commonly used evaluation metrics for image captioning tasks: BLEU\cite{PRWZ2002}, METEOR\cite{BL2005}, ROUGE-L\cite{L2004}, CIDEr\cite{VZP2015}, and SPICE\cite{PBMS2016}. These metrics were calculated with the use of cocoapi\footnote{\href{https://github.com/cocodataset/cocoapi}{https://github.com/cocodataset/cocoapi}}.

\begin{enumerate}
    \item BLEU is a commonly used automatic evaluation metric to measure the similarity between generated and reference texts. It calculates a score by counting the overlapping n-gram phrases (ranging from single words to multiple words) between the generated and reference texts and then combining exact and partial matches.
    \item METEOR combines both exact and non-exact matching criteria such as synonyms and stemming. METEOR uses an external dictionary to help determine synonyms and stemming, providing a more comprehensive assessment of the quality of the generated text.
    \item ROUGE\_L is an evaluation metric used to measure recall between generated and reference texts. It assesses the similarity between generated and reference texts by comparing the length of the longest common subsequence.
    \item CIDEr is a metric used to evaluate image captioning tasks, focusing on the diversity and consensus of the generated descriptions. It calculates scores by comparing word frequency statistics between the generated text and several reference texts, emphasizing the consistency between different reference texts.
    \item SPICE is a metric used for evaluating image caption generation tasks. Unlike traditional n-gram-based metrics, SPICE attempts to capture sparse features in the generated text, such as entities and relationships. It uses semantic parse trees to measure the similarity between generated and reference texts, providing better semantic matching.
\end{enumerate}

\begin{table*}[ht]
\centering
\caption{Quantitative Results of the SOTA Methods on the UCM-caption Dataset.}
\label{tab5-2}
\begin{tabular}{ c | c | c c c c c c c c }
\hline
    &   Params(M)   &   BLEU@1     &   BLEU@2     &   BLEU@3     &   BLEU@4     &   METEOR   &   ROUGE-L   &   CIDEr   &   SPICE \\
\hline
PureT* (2022)\cite{WXS2022}   &220     &   85.73   &   80.20   &   75.62   &   71.29   &   46.86   &   82.01   &   349.00  &   47.94   \\
BLIP-2* (2023)\cite{LLSH2023}   &  188   & 88.04   &   82.23   &   76.89   &   71.84   &   \textbf{47.32}   &   83.36   &   380.31  &   54.78   \\
Word Sentence (2021)\cite{WHZL2021}   &13     &   79.31   &   72.37   &   66.71   &   62.02   &   43.95   &   71.32   &   278.71  &   -   \\
GVFGA+LSGA (2022)\cite{ZZYG2022}  & -    &   83.19   &   76.67   &   71.03   &   65.96   &   44.36   &   78.45   &   332.70  &   48.53 \\
MLCA-Net (2022)\cite{CHXZ2022}    & -    &   82.6   &   77.0    &    71.7    &   66.8    &   43.5    &   77.2    &   324.0   &   47.3    \\
GLCM (2022)\cite{WHZL2022}    &10     &   81.82   &   75.40   &   69.86   &   64.68   &   46.19   &   75.24   &   302.79  &   -   \\
GLCM* (2022)\cite{WHZL2022}    &10    &  65.68   &   59.31   &   53.33   &   48.22   &   36.11   &   59.04   &   189.76  &   30.16   \\
\hline
BITA(Ours) &171    &  \textbf{88.89}  &   \textbf{83.12}  &   \textbf{77.30}  &   \textbf{71.87}  &   46.88  &   \textbf{83.76}  &   \textbf{384.50}   &   \textbf{54.88}  \\
\hline
\multicolumn{9}{l}{* represents our own best re-implemented results.}
\end{tabular}
\end{table*}

\begin{table*}[ht]
\centering
\caption{Quantitative Results of the SOTA Methods on the RSICD Dataset.}
\label{tab5-3}
\begin{tabular}{ c | c | c c c c c c c c }
\hline
    &   Params(M) &   BLEU@1     &   BLEU@2     &   BLEU@3     &   BLEU@4     &   METEOR   &   ROUGE-L   &   CIDEr   &   SPICE \\
\hline
PureT* (2022)\cite{WXS2022}   &220     &   77.05   &   65.75   &   56.63   &   49.19   &   37.66   &   67.41   &   275.56  &   48.87   \\
BLIP-2* (2023)\cite{LLSH2023}    &188  &   76.91   &   66.04   &   57.02   &   49.60   &   40.28   &   69.86   &   294.66  &   53.11   \\
Word Sentence (2021)\cite{WHZL2021}   &13    &   72.40   &   58.61   &   49.33   &   42.50   &   31.97   &   62.60   &   206.29  &   -   \\
GVFGA+LSGA (2022)\cite{ZZYG2022}  & -    &   67.79   &   56.00   &   47.81   &   41.65   &   32.85   &   59.29   &   260.12  &   46.83 \\
MLCA-Net (2022)\cite{CHXZ2022}    & -     &   75.7   &   63.4    &    53.9    &   46.1    &   35.1    &   64.6    &   235.6   &   44.4    \\
GLCM 2022()\cite{WHZL2022}    &10     &   77.67   &   64.92   &   56.42   &   49.37   &   36.27   &   67.79   &   254.91  &   -   \\
GLCM* (2022)\cite{WHZL2022}    &10    &  59.88   &   48.13   &   37.96   &   29.97   &   28.17   &   45.54   &   62.21  &   15.20   \\
\hline
BITA(Ours) &171    &  \textbf{77.38}  &   \textbf{66.54}  &   \textbf{57.65}  &   \textbf{50.36}  &   \textbf{41.99}  &   \textbf{71.74}  &   \textbf{304.53}   &   \textbf{54.79}  \\
\hline
\multicolumn{9}{l}{* represents our own best re-implemented results.}
\end{tabular}
\end{table*}

\begin{table*}[ht]
\centering
\caption{Quantitative Results of the SOTA Methods on the NWPU Dataset.}
\label{tab5-4}
\begin{tabular}{ c | c | c c c c c c c c }
\hline
    &   Params(M) &   BLEU@1     &   BLEU@2     &   BLEU@3     &   BLEU@4     &   METEOR   &   ROUGE-L   &   CIDEr   &   SPICE \\
\hline
PureT* (2022)\cite{WXS2022}   &220    &   \textbf{88.82}   &   80.31   &   73.30   &   67.50   &   42.32   &   75.84   &   195.12  &   27.43   \\
BLIP-2* (2023)\cite{LLSH2023}    &188   &87.78   &   80.24   &   73.51   &   \textbf{67.62}   &   45.01   &   78.44   &   193.86  &   33.28   \\
Word Sentence (2021)\cite{WHZL2021}   &13    &   -   &   -   &   -   &   -   &   -   &   -   &   -  &   -   \\
GVFGA+LSGA (2022)\cite{ZZYG2022}  & -    &   -   &   -   &   -   &   -   &   -   &   -   &   -  &   - \\
MLCA-Net (2022)\cite{CHXZ2022}    & -    &   74.5   &   62.4    &    54.1    &   47.8    &   33.7    &   60.1    &   126.4   &   28.5    \\
GLCM (2022)\cite{WHZL2022}    &10    &   -   &   -   &   -   &   -   &   -   &   -   &   -  &   -   \\
GLCM* (2022)\cite{WHZL2022}    &10   &  55.36   &   42.28   &   33.53   &   27.20   &   27.89   &   50.42   &   127.74  &   32.94   \\
\hline
BITA(Ours) &171   &  88.54  &   \textbf{80.70}  &   \textbf{73.76}  &   67.60  &   \textbf{45.27}  &   \textbf{78.53}  &   \textbf{197.04}   &   \textbf{33.65}  \\
\hline
\multicolumn{9}{l}{* represents our own best re-implemented results.}
\end{tabular}
\end{table*}

\subsection{Baseline methods}
In this section, we evaluate the performance of our proposed method with five SOTA encoder-decoder-based methods and explain our reasons for selecting these advanced methods.

\begin{enumerate}
    \item PureT \cite{WXS2022} utilizes Swin-Transformer to extract grid-level features. A refining encoder captures relationships within grid-level features, then a decoder generates captions word by word. The interaction between visual and language features is enhanced by pooling grid features. The integration of global features through a refining encoder and decoder improves modeling capabilities.
    \item BLIP-2 \cite{LLSH2023} is a visual-language pre-training approach that employs a Q-Former structure inspired by BERT-base for multimodal fusion. It utilizes three pre-training proxy tasks to mitigate the modality gap between image and text: image-text contrastive learning, image-text matching, and image-grounded text generation.
    \item Word Sentence \cite{WHZL2021} is a typical encoder-decoder framework consisting of a word extractor and a sentence generator. This method aims to extract significant words from remote sensing images and form coherent sentences, which decomposes RSIC into word classification and sorting tasks, better aligning with human comprehension.
    \item GVFGA+LSGA \cite{ZZYG2022} provides innovative improvements to the encoder and decoder parts respectively. GVFGA integrates global and local visual features, using an attention gate for enhanced saliency. A linguistic state (LS) alleviates hidden state load by processing textual features, while LS-guided attention (LSGA) refines fused visual-textual features, aided by an attention gate to remove irrelevant information.
    \item MLCA-Net \cite{CHXZ2022} addresses scale inconsistency and category uncertainty in multi-source remote sensing data. Through multi-level and contextual attention modules, MLCA-Net adaptively aggregates features for different scales and latent context exploration. The LSTM-based decoder aligns visual features and semantic descriptors to reduce ambiguity.
    \item GLCM \cite{WHZL2022} is an attention-based global-local captioning model for RSIC. Global features convey overall visual relevance to sentence words, while local features emphasize individual word discrimination. GLCM leverages both types of features for enhanced representation. 
\end{enumerate}

\subsection{Comparative Experimental Results and Analysis}
\subsubsection{Quantitative Comparison}
Table \ref{tab5-2}, \ref{tab5-3}, and \ref{tab5-4} show the quantitative results of various SOTA methods using multiple evaluation metrics on the UCM-caption dataset, the RSICD dataset, and the NWPU-caption dataset, respectively. It's evident that BITA surpasses the other SOTA comparative methods in terms of performance. This is attributed to the capability of the VLP paradigm to leverage the representative and reasoning abilities of both VFMs and LLMs. Crucially, the proposed IFT module plays a vital role in bridging the multimodal gap between VMFs and LLMs. Hence, the proposed approach can align image-text information while generating high-quality image captions.

From the results obtained on the UCM-caption and RSICD datasets, it is evident that methods utilizing CNN as a visual feature extractor and LSTM as a text decoder (such as Word Sentence, GVFGA+LSGA, MLCA-Net, and GLCM) perform noticeably weaker across various metrics compared to methods employing Transformer as both encoder and decoder (such as PureT and BITA). Furthermore, the GVFGA+LSGA approach, which integrates global and local visual features and incorporates a linguistic-state-guided attention mechanism for visual-textual features, exhibits significant performance improvement on RSICD. Additionally, the non-remote sensing method PureT, which introduces interaction between visual and textual features, also outperforms other state-of-the-art comparison methods significantly across various metrics. This underscores the importance of aligning visual-language features. Compared to the second-best method, PureT, the proposed method leverages the strong representation and reasoning capabilities of VFMs and LLMs, together with a dedicated IFT module designed to bridge the gap between visual and textual features. As a result, BITA consistently outperforms PureT across a range of metrics. In particular, our proposed approach demonstrates superior performance in terms of word accuracy and diversity, sentence structure coherence, and semantic consistency. This highlights the effectiveness of the VLP paradigm and the IFT module in feature extraction, text reasoning, and image-text alignment, which together contribute to its impressive results.

Furthermore, a similar phenomenon is evident in the NWPU-caption dataset, where methods combining CNN with LSTM exhibit noticeably weaker performance compared to full Transformer-based approaches. The proposed method continues to lead other comparative methods across all evaluation metrics. However, in contrast to UCM-caption and RSICD, all methods experience a significant drop in evaluation metrics on NWPU-caption. This is because each sentence in the NWPU-caption dataset has a much larger word count than the other two datasets, posing considerable challenges to word accuracy, sentence structure coherence, and semantic consistency in the generated text.

\begin{figure*}
    \centering
    \footnotesize
    \begin{tabular}{p{4cm} p{4cm} p{4cm} p{4cm}}
         \includegraphics[width=4cm]{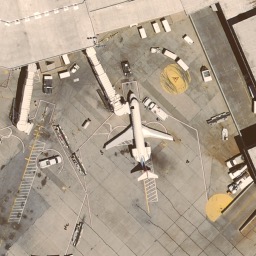} & \includegraphics[width=4cm]{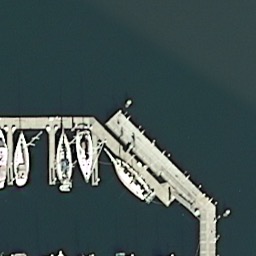} & \includegraphics[width=4cm]{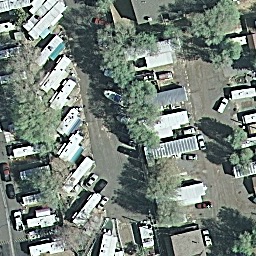}   &   \includegraphics[width=4cm]{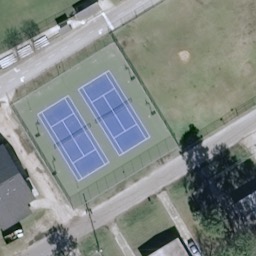}   \\
         \textbf{GT:} An white airplane is stopped at the airport with some luggage cars surrounded it. & \textbf{GT:} Lots of boats docked neatly at the harbor and only a few positions are free. & \textbf{GT:} Many mobile homes arranged neatly in the mobile home park and some roads go through this area.   & \textbf{GT:} There are two tennis courts on the lawn and surrounded by some plants.  \\
         \textbf{GLCM:} There is a big airplane stopped at the airport. & \textbf{GLCM:} There is a piece of a swimming pool beside it. &  \textbf{GLCM:} Lots of mobile homes with plants surrounded in the mobile home park.  &   \textbf{GLCM:} There are two tennis courts arranged neatly and surrounded by some plants.  \\
         \textbf{PureT:} A white airplane is stopped at the airport. & \textbf{PureT:} It is a piece of cropland. &  \textbf{PureT:} Many mobile homes arranged in lines in the mobile home park.   &   \textbf{PureT:} Two tennis courts arranged neatly with a road beside.  \\
         \textbf{BITA:} An airplane is stopped at the airport with some cars beside it. &
         \textbf{BITA:} Lots of boats docked neatly at the harbor and some positions are free. &
            \textbf{BITA:} Many mobile homes arranged haphazardly in the mobile home park and some roads go through this area. &   \textbf{BITA:} There are two tennis courts arranged neatly and surrounded by some plants.\\
            
         \includegraphics[width=4cm]{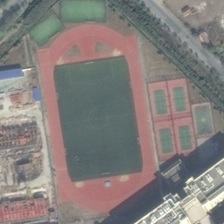}  &   \includegraphics[width=4cm]{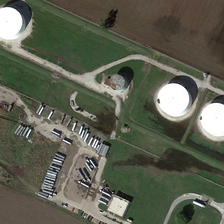} & \includegraphics[width=4cm]{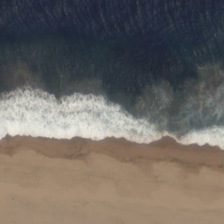} &   \includegraphics[width=4cm]{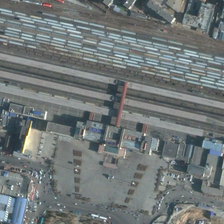} \\
        \textbf{GT:} Some green trees and several buildings are around a playground. & \textbf{GT:} Three storage tanks are surrounded by green meadow. &   \textbf{GT:} White waves in green ocean is near yellow beach.   &   \textbf{GT:} Many buildings are in two sides of a railway station.   \\
        \textbf{GLCM:} Many green trees are around a playground with a row of basketball fields near a playground.  &   \textbf{GLCM:} Three white storage tanks are near two storage tanks with some green trees.  &   \textbf{GLCM:} White waves. &   \textbf{GLCM:} Some buildings and many green trees are in two sides of a railway station.  \\
        \textbf{PureT:} A playground is surrounded by many buildings.   &   \textbf{PureT:} Two white storage tanks are near several buildings and green meadows.   &   \textbf{PureT:} Some green trees are near a piece of yellow beach.  &   \textbf{PureT:} Some buildings are in two sides of a railway station.   \\
        \textbf{BITA:} Many buildings and green trees are around a playground with a basketball field in it.   &   \textbf{BITA:} Several storage tanks are near meadows and green trees. &   \textbf{BITA:} Yellow beach is near a piece of green ocean with white waves.   &   \textbf{BITA:} Many buildings are in two sides of a railway station.   \\

          \includegraphics[width=4cm]{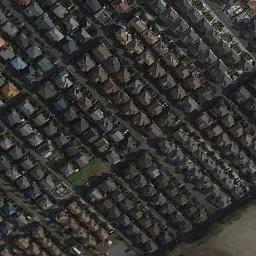} & \includegraphics[width=4cm]{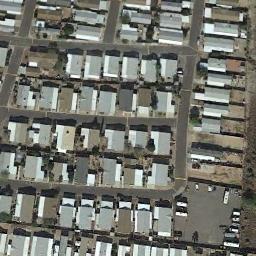}  &   \includegraphics[width=4cm]{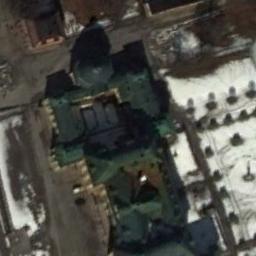} &   \includegraphics[width=4cm]{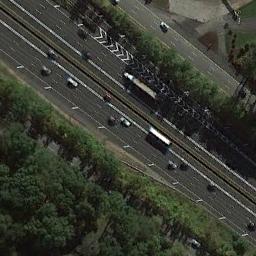} \\
          \textbf{GT:} There are many roads and neatly arranged houses and trees and large lawns in densely populated areas. &
         \textbf{GT:} The dense mobile home park has lots of neatly arranged mobile homes and some roads go through the mobile home park.   &   \textbf{GT:} A black palace with a dome beside a white building.    &   \textbf{GT:} There are large green forests around the freeway and there are many cars on the freeway.   \\
         \textbf{GLCM:} There are many roads of different lengths in a dense residential area. &
         \textbf{GLCM:} There are many buildings in the mobile home park beside the mobile home park.   &   \textbf{GLCM:} There are some green trees and a church beside the church.   &   \textbf{GLCM:} There are green lawns and some bare land beside the freeway.   \\
         \textbf{PureT:} The dense residential is next to the road. &
         \textbf{PureT:} The mobile home park is on the bare land next to the road. &   \textbf{PureT:} The palace is surrounded by trees.  &   \textbf{PureT:} There are green belts along the freeway with many green trees around. \\
         \textbf{BITA:} There are many roads and neatly arranged houses and trees in densely populated areas. &
         \textbf{BITA:} The mobile home park has some neatly arranged white mobile homes and some roads go through the mobile home park.   &   \textbf{BITA:} The church with a cross-shaped roof is on the open place next to some trees.    &   \textbf{BITA:} There are many green trees around the freeway and there are many cars on the freeway.    \\
    \end{tabular}
    \caption{Examples of captions generated by GLCM, PureT, and BITA across three datasets: the results for UCM-caption, RSICD, and NWPU-caption datasets are shown in the first, second, and third rows, respectively. \textbf{GT} represents one of the five ground truth annotations from the original dataset.}
    \label{ch5e-fig-1}
\end{figure*}

\subsubsection{Qualitative Comparison}
Fig. \ref{ch5e-fig-1} displays representative image captions from three experimental datasets, as well as the caption generation results from three methods: GLCM, PureT, and BITA. 

Comparing the caption generation results of GLCM, PureT, and the proposed method, it is evident that the presented method in this article outperforms the other two methods in terms of word accuracy, sentence diversity, and semantic consistency. Specifically, from the first row, second column of Fig. \ref{ch5e-fig-1}, the proposed approach accurately describes the entire image, while the other two methods provide incorrect descriptions. In the bottom row,  the first column of Fig. \ref{ch5e-fig-1}, our proposed method provides a complete description of the scene and uses the same expression as the ground truth, namely the populated area. However, our approach provides fewer descriptions of the colors and quantities of objects in the image. For example, in Fig. \ref{ch5e-fig-1}, there is a lack of description regarding the color of the airplane and the specific description of the number of water tanks. In the description of the beach, our approach accurately portrays the beach and waves, along with their respective colors. Moreover, compared to other contrasting methods, our approach exhibits greater diversity in its descriptions. For mobile home, tennis court, railway station, and freeway, our approach accurately describes the objects in remote sensing images and their contextual information, with the descriptions closely matching the ground truth.

\begin{figure}[!t]
\centering
\includegraphics[width=0.95\linewidth]{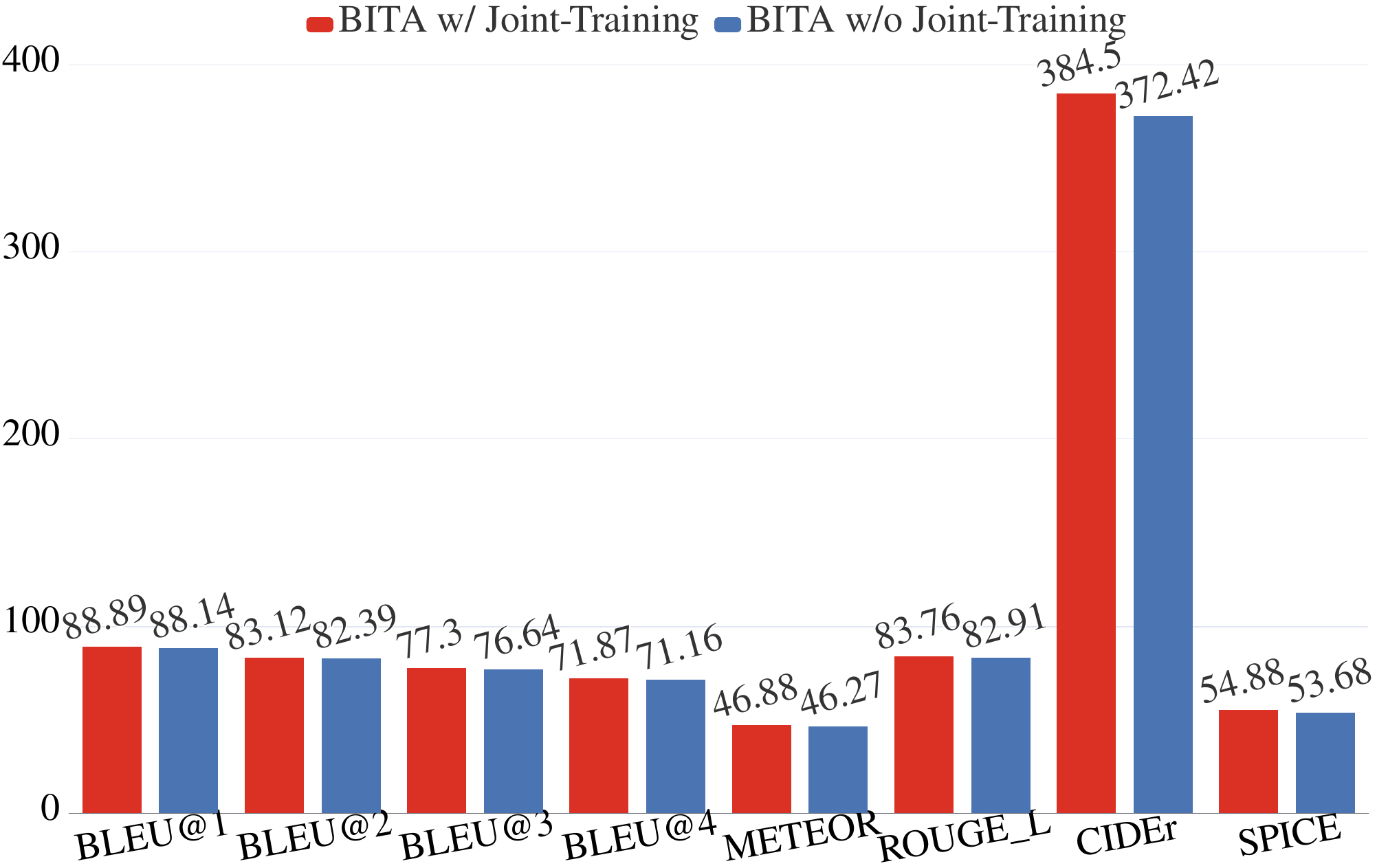}
\caption{Results of whether BITA uses data joint pre-training on UCM-caption dataset.}
\label{ch5f-fig-1}
\end{figure}

\begin{figure}[!t]
\centering
\includegraphics[width=0.95\linewidth]{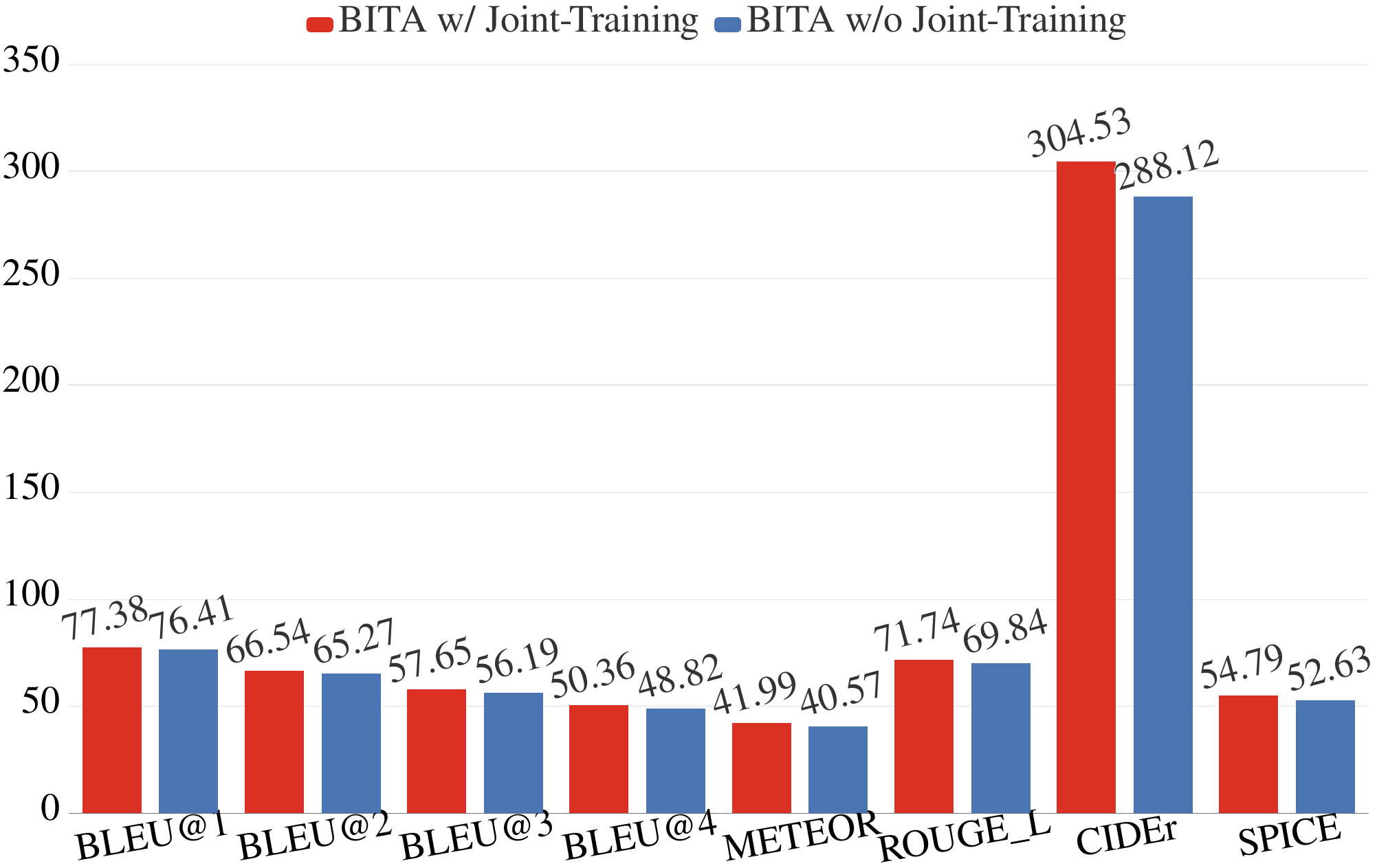}
\caption{Results of whether BITA uses data joint pre-training on RSICD dataset.}
\label{ch5f-fig-2}
\end{figure}

\begin{figure}[!ht]
\centering
\includegraphics[width=0.95\linewidth]{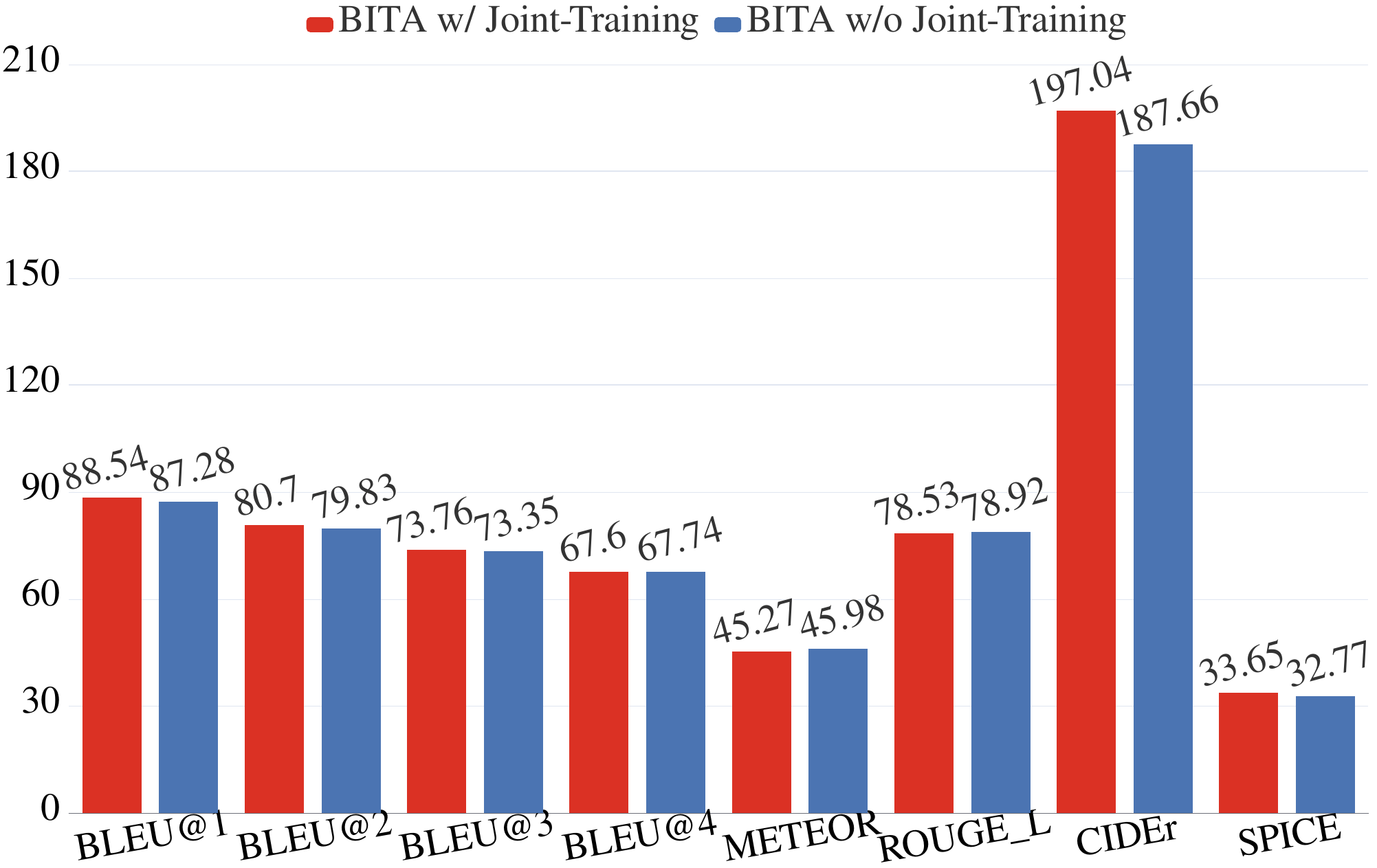}
\caption{Results of whether BITA uses data joint pre-training on NWPU-caption dataset.}
\label{ch5f-fig-3}
\end{figure}

\subsection{Ablation Experiments}
\subsubsection{Impact of Multi-Dataset Joint Pre-Training}
This section primarily emphasizes the importance of conducting pre-training tasks using all training image-text pairs from the three caption datasets, referred to as joint data pre-training. Figs. \ref{ch5f-fig-1}, \ref{ch5f-fig-2}, and \ref{ch5f-fig-3} respectively depict the results of BITA with joint data pre-training and non-joint data pre-training on the UCM-caption, RSICD, and NWPU-caption datasets. Observing the results on the UCM-caption and RSICD datasets, when joint data pre-training is utilized, there is a significant improvement in BITA's performance across metrics such as word accuracy, diversity, sentence coherence, and semantic consistency, especially for datasets with a smaller number of samples. However, when dealing with datasets containing a larger number of samples, joint data pre-training contributes more to enhancing semantic consistency in BITA's generated text, with a relatively weaker improvement in word accuracy. Notably, there is a slight drop in sentence coherence for the generated text of our approach in this scenario.

Taking all aspects into consideration, joint data pre-training has a more positive impact on the performance of BITA with fewer samples, outweighing the performance suppression effect on datasets with a larger sample count. As a result, we select multi-dataset joint pre-training to enhance the model's performance on datasets with fewer samples.


\begin{table*}[!h]
\centering
\caption{Evaluating the Importance of Two-Stage Pre-Training on the RSICD Dataset.}
\label{tab5-5}
\begin{tabular}{ c | c c c c c c c c c c}
\hline
    &   BLEU@1  &   BLEU@2  &   BLEU@3  &   BLEU@4  &   METEOR  &   ROUGE-L &  CIDEr   &   SPICE   \\
\hline
BITA w/o s2    & 75.02   &   63.21   &   53.73   &   46.19   &   39.23   &   68.02   &   277.85  &   50.74   \\
\hline
BITA(Ours)  &  \textbf{77.38}  &   \textbf{66.54}  &   \textbf{57.65}  &   \textbf{50.36}  &   \textbf{41.99}  &   \textbf{71.74}  &  \textbf{304.53}   &   \textbf{54.79}  \\
\hline
\multicolumn{9}{l}{"w/o s2" indicates the absence of the second pre-training stage.}
\end{tabular}
\end{table*}

\begin{table*}[!h]
\centering
\caption{Evaluating the Lightweight and Performance of BITA on the RSICD Dataset.}
\label{tab5-6}
\begin{tabular}{ c | c c | c c c c c c c c}
\hline
    &   Params(M)  &   Time(s)   &   BLEU@1  &   BLEU@2  &   BLEU@3  &   BLEU@4  &   METEOR  &   ROUGE-L &  CIDEr   &   SPICE   \\
\hline
BITA w/o IFT    &  188 &  0.7561  &   76.01   &   66.26   &   57.40   &   50.03   &   39.82   &   70.42   &   298.59  &   52.41   \\
\hline
BITA(Ours)  &  \textbf{171} & \textbf{0.7427}  &  \textbf{77.38}  &   \textbf{66.54}  &   \textbf{57.65}  &   \textbf{50.36}  &   \textbf{41.99}  &   \textbf{71.74}  &  \textbf{304.53}   &   \textbf{54.79}  \\
\hline
\multicolumn{11}{l}{"w/o IFT" represents replacing the IFT module with the encoder from BERT-base.}
\end{tabular}
\end{table*}


\subsubsection{Importance of Two-Stage Pre-Training}
We conduct experiments on the RSICD dataset to validate the importance of the two-stage pre-training. In the first stage of pre-training, we utilize the IFT module and the ITC loss to align image-text representations. The second stage of pre-training involves connecting the frozen visual encoder and the frozen LLM using the IFT module. The process is guided by the PCLM loss to enhance the model's ability to generate text from visual features.

As shown in Table \ref{tab5-5}, if the model does not undergo the second pre-training stage, its performance is significantly compromised, resulting in generated sentences lacking word accuracy, sentence coherence, and semantic consistency. This observation highlights the critical role of the PCLM-guided second-stage pre-training in enhancing the text generation capabilities of the proposed method.

\subsubsection{Speed Performance of IFT}
 Given that the design of the IFT structure is inspired by the BERT-base model, to further validate the effectiveness of IFT, we replaced IFT with the BERT-base structure (BITA w/o IFT). All the model parameters and the pre-training time are measured in the first training stage before the whole pre-training is completed to evaluate the model performance. Compared to using self-attention at each layer in BERT, the IFT module leverages Fourier layers to efficiently extract effective visual features in the frequency domain. It also employs cross-attention layers in every BERT block to extract the most relevant and text-related visual features.

As shown in the experimental results in Table \ref{tab5-6}, BITA has significantly fewer total trainable parameters and consumes less time per iteration compared to "BITA w/o IFT". Most importantly, BITA outperforms "BITA w/o IFT" across all evaluation metrics. This clearly demonstrates that our approach is not only more lightweight than "BITA w/o IFT" but also maintains or even slightly improves performance.

\section{Conclusion}
 In this paper, we have introduced an efficient and powerful two-stage vision-language pre-training method called BITA for remote sensing image captioning tasks. In the first stage, an interactive Fourier Transformer is designed to interact with learnable visual prompts and visual features obtained from the frozen image encoder. This facilitates the capture of the most effective visual features using low-dimensional visual prompts. Additionally, the interactive Fourier Transformer is capable of extracting text features. Subsequently, we utilize image-text contrastive learning to align visual prompts with text features, minimizing the modality gap. In the second stage, the interactive Fourier Transformer serves as a bridge connecting the frozen visual encoder with the LLM, adopting prefix causal language modeling to further enhance the model's ability to generate text from images. Furthermore, when compared to other state-of-the-art methods, the proposed method consistently demonstrates superior performance across three remote sensing image captioning datasets.

Although BITA showcases a combination of lightweight design and high performance, it's worth noting that the two-stage pre-training process can still be intricate. Therefore, future research could explore the design of a single-stage pre-training that maintains both efficiency and performance.

\bibliographystyle{IEEEtran}
\bibliography{Bibtex/ref.bib}

\end{document}